\documentclass{article}

\usepackage{PRIMEarxiv}

\usepackage[utf8]{inputenc} % allow utf-8 input
\usepackage[T1]{fontenc}    % use 8-bit T1 fonts
\usepackage{hyperref}       % hyperlinks
\usepackage{url}            % simple URL typesetting
\usepackage{booktabs}       % professional-quality tables
\usepackage{amsfonts}       % blackboard math symbols
\usepackage{nicefrac}       % compact symbols for 1/2, etc.
\usepackage{microtype}      % microtypography
\usepackage{lipsum}
\usepackage{fancyhdr}       % header
\usepackage{graphicx}       % graphics
\graphicspath{{media/}}     % organize your images and other figures under media/ folder

\title{Object Detection performance variation on compressed satellite image datasets with iquaflow
}

\author{
  Pau Gallés, Katalin Takats and Javier Marin \\
  Satellogic Inc. SATL-NASDAQ  \\
  Barcelona\\
  \texttt{\{pau.galles, katalin.takats, jmarin\}@satellogic.com} \\
}

\begin{document}
\maketitle

\begin{abstract}
Increasing the performance of predictive models on images has been in the focus of many research projects lately.However, studies about the resilience of these models when they are trained on image datasets that suffer modifications altering their original quality are less common, even though their implications are often encountered in the industry \cite{Lofqvist2021},\cite{Jo2021ImpactOI},\cite{Delac2005}. A good example of that is with earth observation satellites that are capturing many images. The energy and time of connection to the earth of an orbiting satellite are limited and must be carefully used. An approach to mitigate that is to compress the images on board before downloading. The compression can be regulated depending on the intended usage of the image and the requirements of this application. We present a new software tool with the name \textsc{iquaflow} that is designed to study image quality and model performance variation given an alteration of the image dataset. Furthermore, we do a showcase study about oriented object detection models adoption on a public image dataset DOTA \cite{Xia_2018_CVPR} given different compression levels. The optimal compression point is found and the usefulness of \textsc{iquaflow} becomes evident.
\end{abstract}

% keywords can be 

\section{Introduction}

Predictive models that use images as inputs are constrained to any image alteration that can degrade the optimal performance of these models. Sometimes the degree of modification on the images can be regulated. A good example is when images are compressed before being sent to the algorithm for prediction. In case of earth observation satellites, the high cost of downloading the images can be significantly reduced by compressing the images first. \cite{Lofqvist2021}. One approach is to make images smaller to reduce the costs of downloading to earth \cite{rs13030447}. In this context, decision-makers need tools to study the optimal modification so that the performance of the predictive models is adequate despite the compression. \textsc{iquaflow}\footnote{\href{https://github.com/satellogic/iquaflow}{https://github.com/satellogic/iquaflow}} is a software tool that has been designed precisely to study image quality as well as the performance of models trained on top of provided datasets that are modified with any user-defined alteration. \cite{Lofqvist2021} studies  object detection inference with compression algorithms based on decimation and scaling with interpolation in the context of earth observation from satellite applications. In the present work, the study is brought further with custom training for each level of compression, new kinds of compression, and new models of object detection that are suitable for oriented annotations as explained below.

\subsection{Compression}

Compression algorithms can be lossless or lossy \cite{Hussain2018ImageCT}. The first kind performs an operation on the image that allows the recovery of the original image before it was compressed. The second kind, on the other hand, does an irreversible operation. Using a lossy compression algorithm, we can achieve a greater reduction in file sizes than with a lossless one. A simple straightforward technique for lossy compression can be the interpolation of an image to fewer pixels. Then a smaller image will have lost information and it will also be smaller in file size. In this study the JPEG compression is used as explained in section \ref{compresssect}.

\subsection{Object detection} 

A good example of predictive models on images is object detection (such as vehicles from aerial images). Most detectors such as Faster R-CNN \cite{renNIPS15fasterrcnn}, SSD \cite{LiuAESRFB16} and YOLOv2, v3 \cite{redmon2018yolov3} rely on a set of pre-defined anchors that consist in a small set of bounding boxes summarizing the most relevant geometric shapes covering relevant scale, aspect ratios and elongation directions. The idea is that any object can be associated with a specific anchor box without having to have a perfect fit. 

However, the definition of this set of anchor boxes is a hyper-parameter that must be defined and has an effect on the detection performance. The models are, of course, sensitive to the sizes, aspect ratios, and a number of anchors defined in the set (see \cite{renNIPS15fasterrcnn} and \cite{lin2017focal}).

Another aspect to consider is the number of stages. Detectors can be composed of multiple stages and each of them has a trained model that solves a specific task in the workflow. A typical case in an object detection problem is the Region Proposal Network which is responsible for the task of generating bounding box proposals. Examples of that are \cite{DBLP2018}, \cite{Xu_2021} and \cite{SCRDet2018}. One advantage of the multistage approach is that each step in the workflow can be easily defined and understood by human logic. In single-stage detectors, the logic can be difficult to interpret inside an end-to-end network solution.

Depending on the annotations one can use a model that predicts with horizontal bounding boxes (HBB) or oriented bounding boxes (OBB). One problem with HBB is distinguishing between overlapping instances of different objects. This is usually approached with the logic of Non-Maximum Suppression (NMS) that involves the measure of Intersection Over Union between different instances to asses the overlapping and whether or not candidate boxes belong to the same sample. This logic struggles when there are elongated objects that are diagonal and parallel to each other. In aerial images, these can be ships in a harbor or trucks in parking. One solution for this is to consider more complex geometries that have a better fit with the object. The simplest complexity, in this case, is to orient the bounding box.

The models used in this study are explained in section \ref{objectdectsect}.

\subsection{Iquaflow}

Image quality can be often evaluated by the human eye. However, it is very challenging to define a numerical measurement for image quality. One of the reasons is that there are many aspects to consider such as the blur, the noise, the quality distribution along frequencies, etc. Moreover, image quality should be measured according to the particular application of the images being measured.  Supervised super-resolution image prediction models are algorithms that translate an input image to a higher-resolution image that contains more pixels. These models are trained with a database containing pair samples of images with their respective higher resolution (also known as ground-truth or target images). In this context, the evaluation of quality will perform better by comparing the predicted image against the target image. These metrics are also known as similarity metrics and they include \cite{uSSIM2020}, \cite{DeepPerc2018}, \cite{Reisenhofer_2018} and \cite{Ding_2020}. Another context is when images are used as inputs for other predictive models with the aim to collect information from them. It is the case of an image classifier or object detection. For this case, a suitable image quality evaluation method can be the performance of this model on the images. This is assuming that changes in the input image quality are affecting the performance of the prediction model. Again, this is a way to measure image quality that is adapted to the actual application of the image. 

\textsc{iquaflow}\cite{iquaflow} is a python package tool that measures image quality by using different approaches. Deterministic metrics include blind metrics which are measured directly on the image without comparing against a reference image or similarity metrics when they are measuring affinity against an ideal case. There are two metrics that have been designed for \textsc{iquaflow} which are implicit measurements of blur and noise levels. The first relies on edges found within the images and it measures the slope in the Relative Edge Response (RER) \cite{RER}. Then the second is based on homogeneous areas where the noise can be estimated (Signal to Noise ratio - SNR) \cite{SNR}. The Quality Metric Regression Network (QMRNet)\cite{qmrnet} has been designed, trained, and integrated into \textsc{iquaflow}. This is a classifier with small intervals that can predict quality parameters on images such as blur (measured as equivalent sigma from a gaussian bell), sharpness, pixel size and noise. Quality can also be measured by checking how predictive models trained on the image dataset are performing. A good example is the present study where object detection is trained on different quality datasets with different outcomes.

Apart from measuring image quality, \textsc{iquaflow} has a whole ecosystem that facilitates the design of new studies and experiment sets made of several training runs with variations. \textsc{iquaflow} wraps another open source tool named \href{https://mlflow.org/}{Mlflow} that is used for machine learning experiment tracking. It will record the executions in a standard format so that they are later easily visualized and compared from \href{https://mlflow.org/}{Mlflow} user interface tool in the browser. In \textsc{iquaflow} the user can add custom metrics and dataset modifiers that are easily integrated into a use case study.
%%%%%%%%%%%%%%%%%%%%%%%%%%%%%%%%%%%%%%%%%%
\section{Materials and Methods}

The aim of the study is to measure the variation of the object detection algorithm's performance on a given image dataset that is modified with various compression ratios. Our goal is to evaluate what is the maximum compression level that still allows for acceptable model performance. In this section, the compression algorithm is described, and the object detection model(s) that we considered, as well as the tool used for managing our experiments.

\subsection{Data} 

Two different datasets are used to carry out two experiments. The first analysis is based on the airplanes dataset\footnote{Contact iquaflow@satellogic.com to request access to the dataset} which consists of 998 images of 1024 × 1024 pixels from airport areas with a total of almost $17000$ annotated planes. These captures were made using NewSat Satellogic constellation ($1~\mathrm{m}$ GSD) and the annotations were made using Happyrobot\footnote{\href{https://happyrobot.ai}{https://happyrobot.ai}} platform. The training partition contained $13731$ annotations and the remaining were used for evaluation.

The second experiment was based on the public dataset DOTA \cite{Xia_2018_CVPR}. It is a dataset for object detection in aerial images. The images are collected from different sensors the image sizes are ranging from $800~\times~800$ to $20000~\times~20000$ pixels and the pixel size varies from $0.3~\mathrm{m}$ to $2~\mathrm{m}$ resolution. DOTA has several versions and DOTA-v1.0 has been used in the present study which contains 15 common categories \footnote{plane, ship, storage tank, baseball diamond, tennis court, basketball court, ground track field, harbor, bridge, large vehicle, small vehicle, helicopter, roundabout, soccer ball field and swimming pool.}, $2806$ images and more than $188k$ object instances. The annotations are oriented bounding boxes which allows us to train both oriented (OBB) and horizontal bounding boxes (HBB) models. The proportions of the training set, validation set, and testing set in DOTA-v1.0 are $1/2$, $1/6$, and $1/3$ \cite{Xia_2018_CVPR}. A disadvantage of this dataset is that the test set is not openly available, rather it is in a form of a remote service to query the predictions. This does not allow to alter the test on the same way the other partitions are modified in the present study. Because of that, 2 partitions are made from the validation set: half of is is used as actual validation and the other half for testing. Then the images are cropped to $1024~\times~1024$ with padding when necessary. After this operation the number of crops for the partitions train, validation and testing are respectively $9734$, $2670$ and $2627$.

\subsection{Compression} \label{compresssect}

In this study, JPEG compression \cite{Wallace91} is used. It is a lossy form of compression based on the discrete cosine transform (DCT) that converts images into the frequency domain and discards high-frequency information by a quantization process. The degree of compression in JPEG can be adjusted: the greater the quality the bigger the file size. In the present study, the compression is set at different levels with the aim to find an optimal value with respect to the performance of predictive models trained on them. We used the JPEG compression from OpenCV \cite{opencv_library} that can be regulated with the parameter CV\_IMWRITE\_JPEG\_QUALITY which can vary from $0$ to $100$ (the higher is the better) with a default value of 95.

\begin{figure}
\centering
\includegraphics[width=10.5 cm]{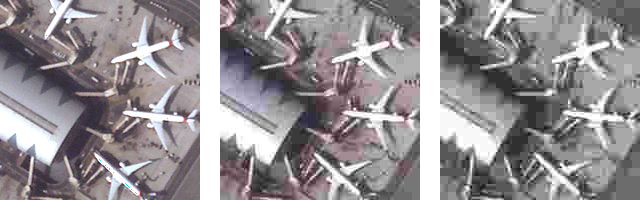}
\caption{JPEG compression effects (original, JPG10,and JPG5 from left to right). This image is from the airplane dataset from Satellogic.
\label{figCompression}}
\end{figure}

Figure~\ref{figCompression} shows an example of the effect when compressing one of the images with JPEG method.

\subsection{Object detection} \label{objectdectsect}

The first experiment has HBB annotated objects and the model YOLOv5 \cite{glenn_jocher_2021_4679653} was used because of its fast training and implementation.

For the second experiment, two OBB models were used. The first was Oriented R-CNN which is a two-stage oriented detector that uses Region Proposal Network (oriented RPN) in the first stage in order to generate high-quality oriented proposals in a nearly cost-free manner \cite{Xie_2021_ICCV}. 

Then the other model used was FCOS \cite{tian2021fcos} which is originally designed for horizontal bounding boxes but it can be adapted with an added convolution layer channel on the top of the regression features that define the direction of the bounding box. Intersection Over Union is often used as a loss function in object detection. However, the IoU calculation between oriented boxes is complex and often not differentiable. There are rotated IoU that implements differentiable IoU calculation for oriented bounding boxes. In this case, the PolyIoULoss \cite{PIoU2021} between the OBB predictions and ground truths is used as a bounding box loss.

The performance of the detector is measured by calculating the average recall (AR) as well as the Mean Average Precision (mAP). AR is a ratio of correctly detected instances over the actual amount of objects. On the other hand, AP is defined with the same correctly detected instances over all the amount of detected cases (including wrong detection). The predicted bounding boxes do not have to have a perfect match with the ground truth. Because of that, the Intersection over Union (IoU) for each prediction and ground truth match candidate is measured to evaluate if they match. In which case it is considered a correct detection \cite{powers2011evaluation}. In this study, mAP is calculated by taking the mean AP over all classes and over a range of IoU thresholds. 

\subsection{Experiment management}

The present study involves a workflow with multiple versions of the original dataset with the corresponding partitions for each altered version (train, validation and test) as well as many training experiment executions and tracking of results that must be organized correctly. All this can be managed easily with a typical \textsc{iquaflow} workflow as follows:

\begin{enumerate}
\item Optionally the user can start with a repository template of \textsc{iquaflow} use cases. This repository uses cookiecutter which is a python package tool for repository templates. By using this you can initialize a repository with the typical required files for a study in \textsc{iquaflow}.
\item The first step will be to set the modifications of the original dataset with different compression levels. This can be done with a list of Modifiers in \textsc{iquaflow}. There are some modifiers already available in \textsc{iquaflow} with performing specific alterations. However, one can set up a custom modifier by inheriting the DSModifier class of \textsc{iquaflow}. The list of modifiers will then be passed as an argument to the experiment setup.
\item Next step is to adapt the user training script to the \textsc{iquaflow} conventions. This is just to accept some input arguments such as the output path where the results are written. Optionally one can monitor in streaming the training by inputting additional arguments as explained in \textsc{iquaflow}'s guide.
\item All previous definitions are introduced in the experimental setup that can be executed afterward. The whole experiment will contain all runs which are the result of combining dataset modifications (the diverse compression levels) and the two different detectors that are used which will be defined as hyperparameter variations in the experiment setup.
\item The evaluation can be either done within the user's custom training script or by using a Metric in \textsc{iquaflow}. Similar to Modifiers there are some specific Metrics already defined in \textsc{iquaflow}. Alternatively, the user can make a custom metric by inheriting the class Metric from \textsc{iquaflow}. The results can be collected from \textsc{iquaflow} or directly by raising an \href{https://mlflow.org/}{mlflow} server which is a tool that is wrapped and used by \textsc{iquaflow}.
\end{enumerate}

As you can see using \textsc{iquaflow} we can automate the compression algorithm on the data, run the user custom training script and evaluate a model. All the results are logged using mlflow and can be handily compared and visualized. \textsc{iquaflow} is the ideal tool for this purpose.

%%%%%%%%%%%%%%%%%%%%%%%%%%%%%%%%%%%%%%%%%%
\section{Results}

\begin{figure}
\centering
\includegraphics[width=10.5 cm]{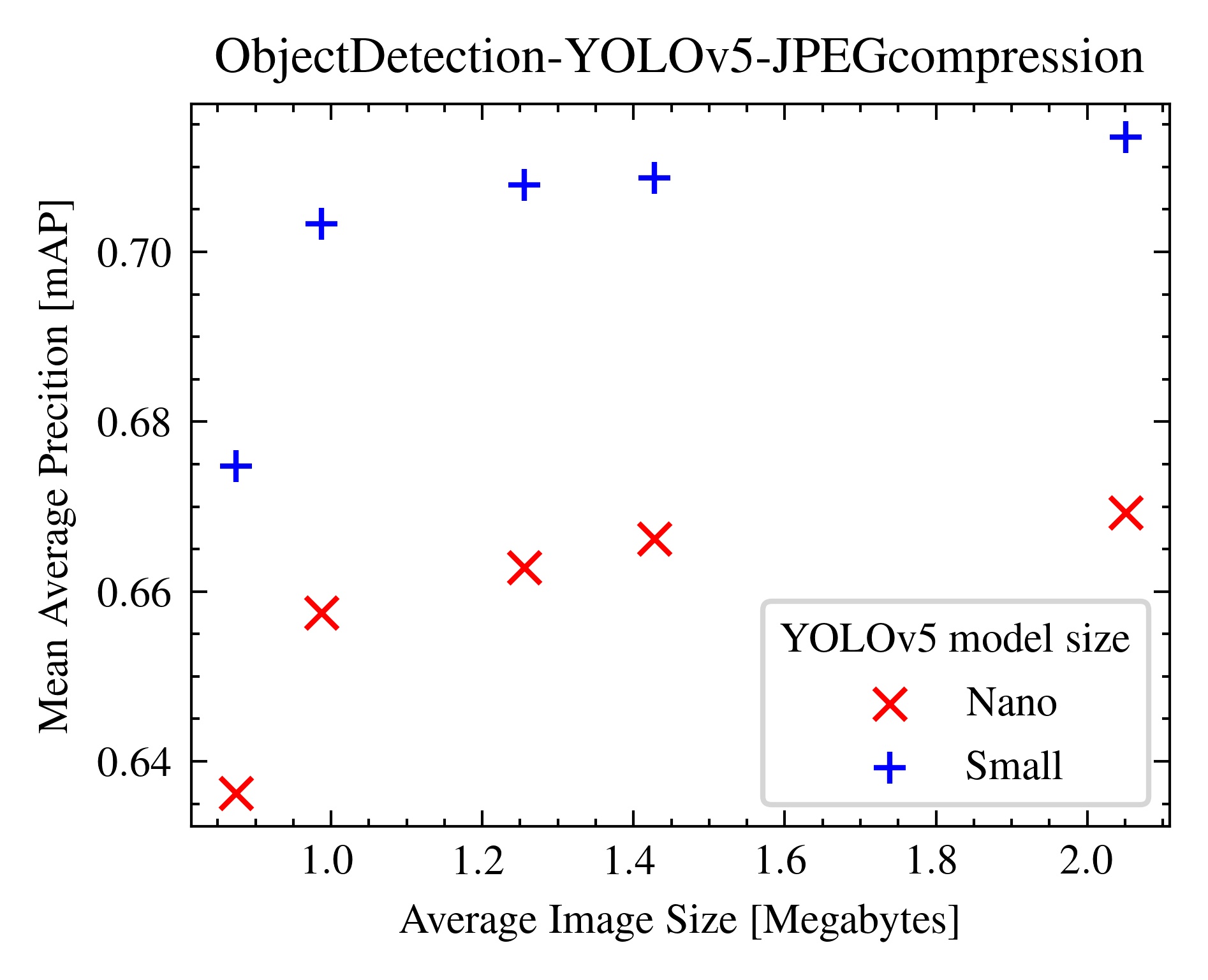}
\caption{Scatter plot that shows the performance of the models (mAP) evolution with different compression levels expressed as average image size of the files in the modified Satellogic's airplanes dataset. Red with "x" and blue with "+" correspond to model size nano and small of YOLOv5 model respectively.
\label{fig1}}
\end{figure}

The airplanes dataset from Satellogic\footnote{\href{https://github.com/satellogic/iquaflow-airport-use-case}{https://github.com/satellogic/iquaflow-airport-use-case}} has the unique category of planes. The image format is tiff and the original average image size is $3.204$Megabytes. The average recall (AR) is measured and the Mean Average Precision (mAP) is calculated over different Intersection Over Union (IoU) thresholds varying from $0.5$ to $0.95$ with a step of $0.05$ and average again for the final score. Table~\ref{table:tab1} contains the resultant metrics and Figure~\ref{fig1} shows performances (mAP) along different levels of compression.

\begin{table}
 \caption{Performance results at different compression levels using the airplanes dataset and two YOLOv5 model sizes with different architecture complexities. The scores for the different models are expressed as Mean Average Precision (mAP) and Average Recall (AR) as expressed in the methodology section. The last column shows the equivalent average image size from the dataset given the level of compression used.\label{tab1}}
  \centering
  \begin{tabular}{lllll}
    \toprule
    \multicolumn{2}{c}{\textbf{YOLOv5 NANO}} & \multicolumn{2}{c}{\textbf{YOLOv5 SMALL}} & size \\
    \cmidrule(r){1-5}
    \textbf{AR}	& \textbf{mAP}	& \textbf{AR} & \textbf{mAP} & \textbf{Mb} \\
    \midrule
    0.898 & \textbf{0.669} & \textbf{0.922} & \textbf{0.714} & 2.051 \\
    \textbf{0.899} & 0.666 & 0.919 & 0.709 & 1.428 \\
    0.892 & 0.663 & 0.917 & 0.708 & 1.256 \\
    0.888 & 0.657 & 0.916 & 0.703 & 0.988 \\
    0.872 & 0.636 & 0.891 & 0.675 & 0.874 \\
    \bottomrule
  \end{tabular}
  \label{table:tab1}
\end{table}

\begin{figure}
\centering
\includegraphics[width=10.5 cm]{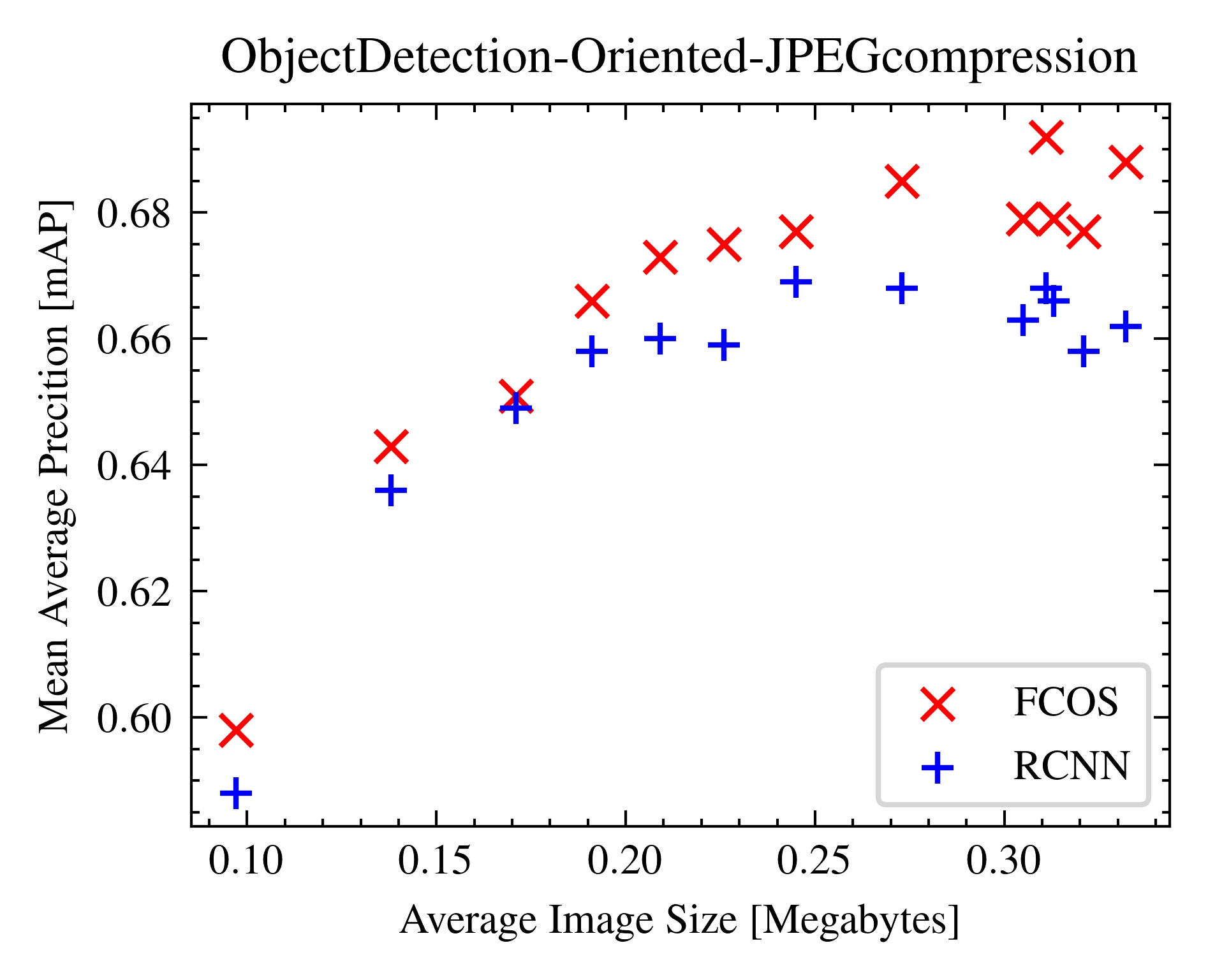}
\caption{Scatter plot that shows the performance of the models (mAP) evolution with different compression levels expressed as average image size of the files using the DOTA1.0 dataset and two OBB models. Red dots correspond to the adapted FCOS model whereas blue dots are from the oriented RCNN model.
\label{fig2}}
\end{figure}

The DOTAv1.0 dataset has 15 categories and different metrics are measured for each class. The categories of 'plane' and 'storage tank' are performing the best whereas the categories 'bridge' and 'soccer-ball-field' are performing the worst. Table~\ref{table:tab2} summarizes the averaged metrics for each run by aggregating with the mean of all the categories. Following the same logic, Figure~\ref{fig2} charts the evolution of performance (mAP) along different levels of compression. The original average image size of the $1024~\times~1024$ crops without compression was $1.13$Megabytes\footnote{\href{https://github.com/satellogic/iquaflow-dota-obb-use-case}{https://github.com/satellogic/iquaflow-dota-obb-use-case}}.

\begin{table}
\caption{Performance results at different compression levels using the DOTA1.0 dataset and two OBB models. The scores for the different models are expressed as Mean Average Precision (mAP) and Average Recall (AR) as expressed in the methodology section. The last column shows the equivalent average image size from the dataset given the level of compression used.\label{tab2}}
\centering
 \begin{tabular}{lllll}
    \toprule
    \multicolumn{2}{c}{\textbf{FCOS}} & \multicolumn{2}{c}{\textbf{RCNN}} & size \\
    \cmidrule(r){1-5}
    \textbf{AR}	& \textbf{mAP}	& \textbf{AR} & \textbf{mAP} & \textbf{Mb} \\
    \midrule
    \textbf{0.869} & 0.688 & 0.806 & 0.662 & 0.332 \\
    0.856 & 0.677 & 0.812 & 0.658 & 0.321 \\
    0.865 & \textbf{0.692} & \textbf{0.813} & \textbf{0.668} & 0.311 \\
    0.861 & 0.679 & 0.812 & 0.666 & 0.313 \\
    0.861 & 0.679 & 0.810 & 0.663 & 0.305 \\
    0.861 & 0.685 & 0.806 & 0.668 & 0.273 \\
    0.849 & 0.677 & 0.811 & 0.669 & 0.245 \\
    0.856 & 0.675 & 0.804 & 0.659 & 0.226 \\
    0.847 & 0.673 & 0.800 & 0.660 & 0.209 \\
    0.846 & 0.666 & 0.798 & 0.658 & 0.191 \\
    0.835 & 0.651 & 0.785 & 0.649 & 0.171 \\
    0.831 & 0.643 & 0.785 & 0.636 & 0.138 \\
    0.799 & 0.598 & 0.741 & 0.588 & \textbf{0.097} \\
    \bottomrule
\end{tabular}
\label{table:tab2}
\end{table}

The optimal compression ratio for the oriented-RCNN model seems to be around JPEG quality score of 70 which corresponds to an average image size of 0.245 Megabytes. This is because it corresponds to the minimum average file size that can be defined without lowering the performance. On the other hand, the adapted FCOS model seems to have an optimal around 80 for JPEG quality score which corresponds to an average image size of 0.273 Megabytes.

\unskip

\begin{figure}[ht]
\centering
\includegraphics[width=10.5 cm]{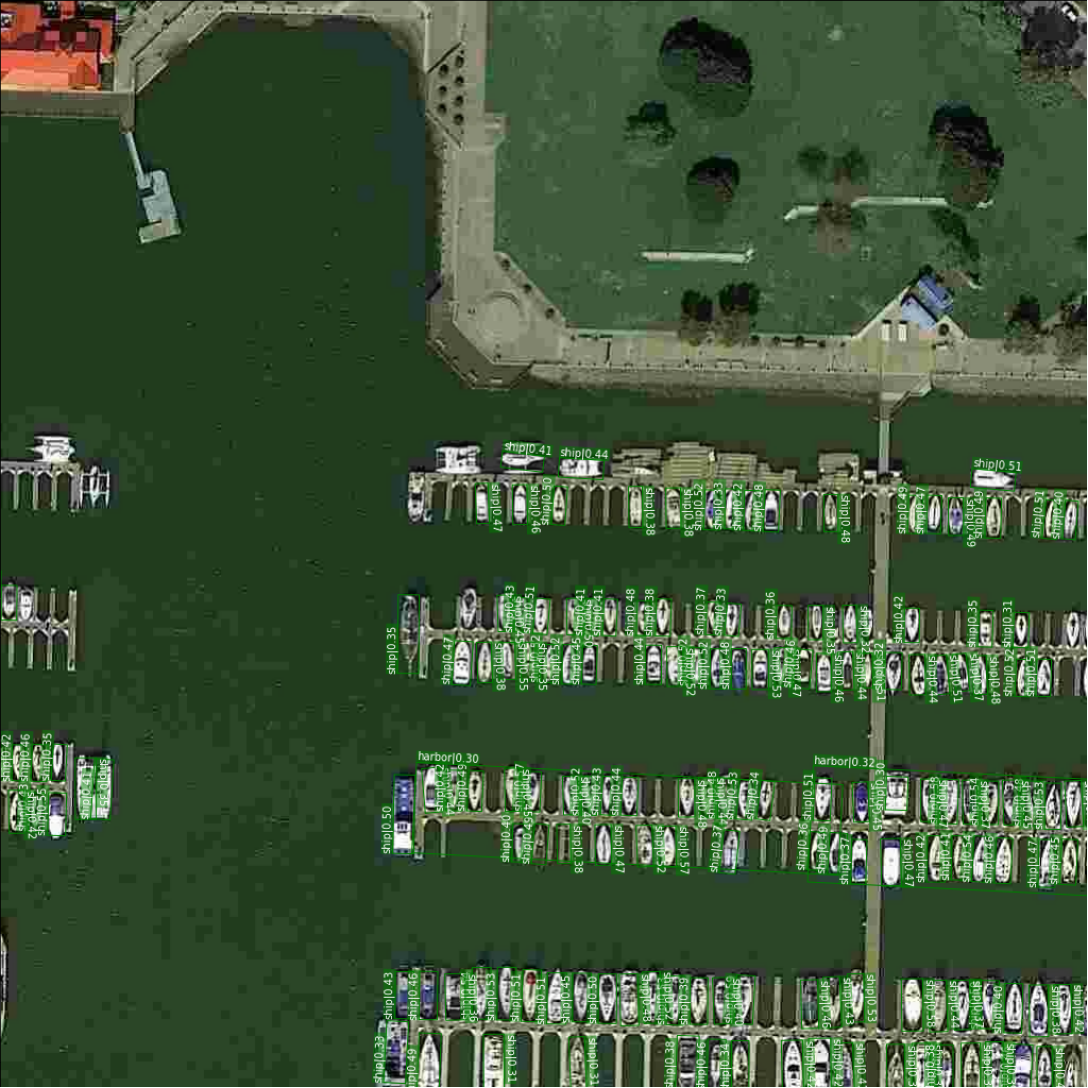}
\caption{An example of prediction on an image with boats compressed with $CV\_JPEG\_QUALITY$ of $10$ which is equivalent to an average dataset image size of $0.097Mb$. The model used is adapted FCOS. The image belongs to the testing partition.
\label{figships}}
\end{figure}

\begin{figure}[ht]
\centering
\includegraphics[width=10.5 cm]{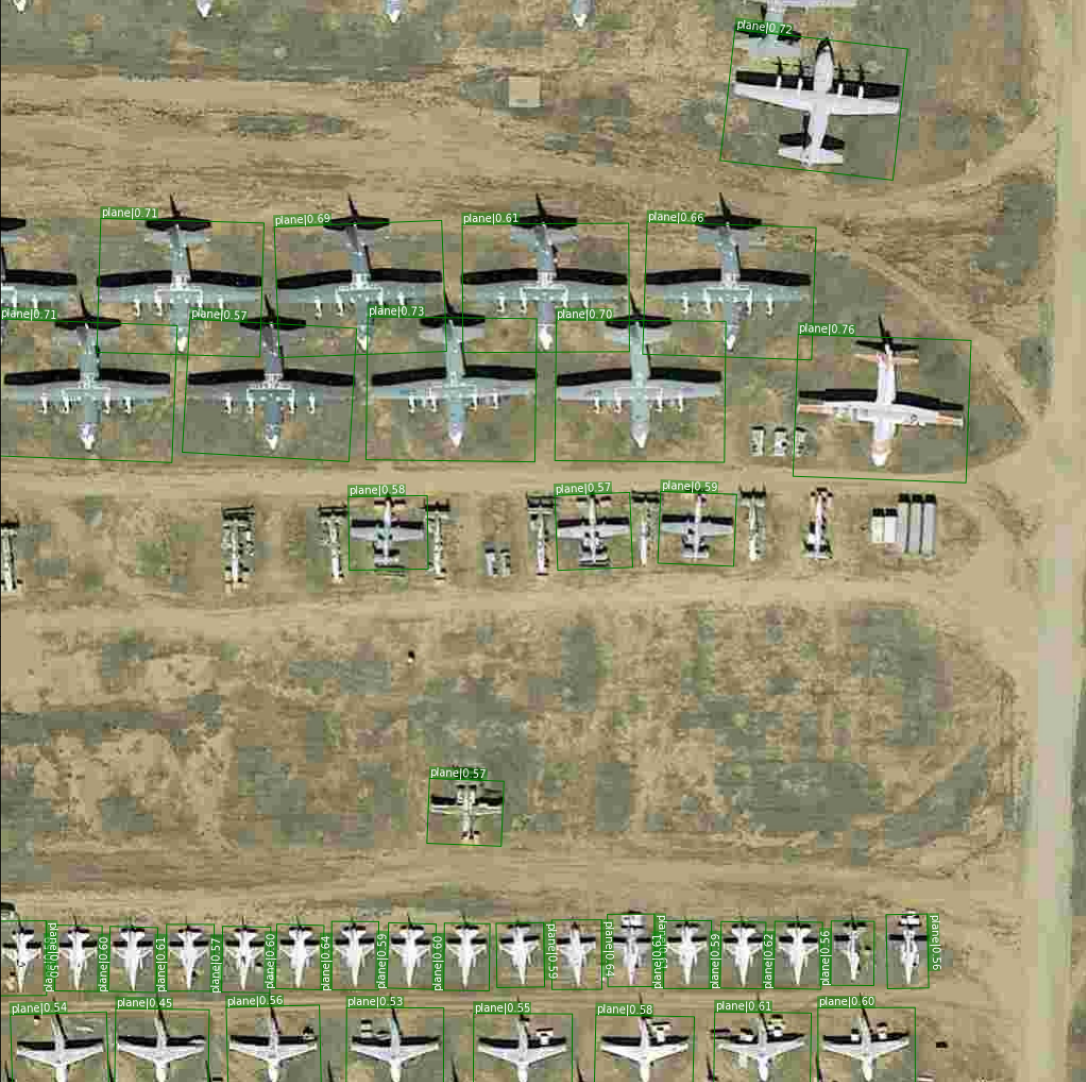}
\caption{An example of prediction on an image with planes compressed with $CV\_JPEG\_QUALITY$ of $10$ which is equivalent to an average dataset image size of $0.097Mb$. The model used is adapted FCOS. The image belongs to the testing partition.
\label{figplanes}}
\end{figure}

%\begin{listing}
%\caption{Title of the listing}
%\rule{\columnwidth}{1pt}
%\raggedright Text of the listing. In font size footnotesize, small, or normalsize. Preferred format: left aligned and single spaced. Preferred border format: top border line and bottom border line.
%\rule{\columnwidth}{1pt}
%\end{listing}

%%%%%%%%%%%%%%%%%%%%%%%%%%%%%%%%%%%%%%%%%%
\section{Conclusions}

In the experiment with Satellogic's airplanes dataset, the decrease in performance with compression is consistent for both models. The variations of mAP is small between the ranges of $0.15$ and $0.25$ average image size. The additional complexity of the Small model has a constant positive shift of $0.5$ in mAP with respect to the Nano model along all the analyzed compression rates.

In the context of the second experiment the adapted FCOS model seems to perform better than oriented RCNN because the AR and mAP are greater for all levels of compression. On the other hand, oriented-RCNN seems more resilient because the optimal compression ratio is higher than the optimal case for the other model. However, the degraded performance of FCOS model given the same compression setting as the optimal value for oriented-RCNN still offers higher performance. FCOS is also easier to implement because it is a single-stage detector that does not require setting anchors as hyperparameters. So far, given the data and context of the study, FCOS seems the best option.

Another interesting observation is the high resilience of the model for some specific applications. The figures \ref{figships} and \ref{figplanes} show a prediction with the FCOS model on an image with boats and airplanes respectively. Both of the images were set with a compression rate of $10$ for $CV\_JPEG\_QUALITY$ which is equivalent to an average dataset image size of $0.097Mb$. In the first image, $146$ ships were correctly detected (True positives), $9$ were wrongly detected (False positive) and $11$ ships were missed (False negative). In the other example all the planes (total amount: $39$) are correctly detected see \ref{figplanes} with no false positives or false negatives. This highlights the greater capacity of compressing images for usage such as the detection of airplanes over smaller or more difficult objects.

This study highlights the potential of \textsc{iquaflow} for decision-makers as well as researchers that want to study performance variation in an agile and ordered way. The key effort has been the development of the tool so that it facilitates further studies with the aim to scale it. The tool also allows for mitigating the uncertainty of image quality by using several strategies to measure that. This is helping also in studies that are exploring suitable solutions for satellite image Super Resolution.

\section*{Acknowledgments}
Conceptualization, P.G. and J.M.; methodology, P.G. and J.M.; software, P.G. and K.T.; validation, K.T. and J.M.; formal analysis, P.G.; investigation, P.G.; resources, J.M.; data curation, P.G.; writing---original draft preparation, P.G.; writing---review and editing, K.T. and J.M.; visualization, P.G.; supervision, J.M.; project administration, J.M.; funding acquisition, J.M. All authors have read and agreed to the published version of the manuscript. 

This research was funded by the Ministry of Science and Innovation and by the European Union within the framework of Retos-Collaboration of the State Program of Research, Development and Innovation Oriented to the Challenges of Society, within the State Research Plan Scientific and Technical and Innovation 2017-2020, with the main objective of promoting technological development, innovation, and quality research. grant number: RTC2019-007434-7. 

The authors declare no conflict of interest.

%Bibliography
\bibliographystyle{unsrt}
\bibliography{references}
\end{document}